\documentclass[runningheads,a4paper]{llncs}

\usepackage{amsmath}
\usepackage{amssymb}
\usepackage{graphicx}
\usepackage[table,pdftex]{xcolor}
\usepackage{xspace}
\usepackage[colorlinks=true,citecolor=blue,urlcolor=black,linkcolor=black]{hyperref}
\usepackage{lineno}
\usepackage{multirow}
\usepackage{subfig}
\usepackage[colorinlistoftodos]{todonotes}
\usepackage[graphicx]{realboxes}
\usepackage{cite}

\usepackage{booktabs} 
\usepackage{makecell} 
\usepackage{arydshln} 
\usepackage[acronym]{glossaries}

\newcommand{\STAB}[1]{\begin{tabular}{@{}c@{}}#1\end{tabular}}

\usepackage{pifont}
\newacronym{soi}{SoI}{Structures-of-Interest}
\newacronym{istn}{ISTNs}{Image-and-Spatial Transformer Networks}
\newacronym{stn}{STNs}{Spatial Transformer Networks}

\begin{document}

\mainmatter  
\title{Image-and-Spatial Transformer Networks\\for Structure-Guided Image Registration}
\titlerunning{Image-and-Spatial Transformer Networks}

\author{Matthew C.H. Lee, Ozan Oktay, Andreas Schuh\\Michiel Schaap$^{*}$, \and Ben Glocker$^{*}$}

\institute{HeartFlow, USA\\Biomedical Image Analysis Group, Imperial College London, UK\\ $^*$ Shared senior authorship.}

\authorrunning{M.C.H. Lee, O. Oktay, A. Schuh, M. Schaap \and B. Glocker}


\maketitle


\begin{abstract}
Image registration with deep neural networks has become an active field of research and exciting avenue for a long standing problem in medical imaging. The goal is to learn a complex function that maps the appearance of input image pairs to parameters of a spatial transformation in order to align corresponding anatomical structures. We argue and show that the current direct, non-iterative approaches are sub-optimal, in particular if we seek accurate alignment of \acrfull{soi}. Information about \acrshort{soi} is often available at training time, for example, in form of segmentations or landmarks. We introduce a novel, generic framework, \acrfull{istn}, to leverage \acrshort{soi} information allowing us to learn new image representations that are optimised for the downstream registration task. Thanks to these representations we can employ a test-specific, iterative refinement over the transformation parameters which yields highly accurate registration even with very limited training data. Performance is demonstrated on  pairwise 3D brain registration and illustrative synthetic data.

\end{abstract}


\section{Introduction}

Image registration remains a fundamental problem in medical image computing, where the goal is to estimate a spatial transformation $T_\theta: \mathbb{R}^d \rightarrow \mathbb{R}^d$ mapping corresponding anatomical locations between $d$-dimensional images. The most widely used approach is intensity-based registration formalised as an optimisation problem seeking optimal transformation parameters $\theta$ that minimise a dissimilarity measure (or cost function) $\mathcal{L}(M\circ T_{\theta}, F)$, where $M$ is the moving source image undergoing spatial transformation and $F$ is the fixed target image. We refer the reader to \cite{sotiras2013deformable} for a detailed overview of what we here call \emph{traditional} methods, i.e., non-learning based approaches making use of iterative optimisation strategies to minimise the cost function for a given pair of images.

Recently, the use of neural networks to learn the complex mapping from image appearance to spatial transformation has become an active field of research~\cite{yang2017quicksilver,sokooti2017nonrigid,rohe2017svf,hu2018weakly,balakrishnan2019voxelmorph,deVos2019deep}, providing a new perspective on tackling challenging registration problems. So called \emph{supervised} approaches \cite{yang2017quicksilver,sokooti2017nonrigid} have been used, which are similar in nature to methods used for image segmentation, where a convolutional neural network is trained to predict the transformation directly using examples of images and their ground truth transformation. Because actual ground truth for $T_{\theta}$ is not available, either random transformations are used to generate synthetic examples or a well established traditional registration method is employed to obtain reference transformations. Neither is optimal, as synthetic transformations might not be realistic and/or yield poor generalisation, while when employing a traditional method the prediction accuracy of the trained network is inherently limited by the accuracy of that method. One might argue that in this case the neural network is mostly learning to replicate the traditional method, although with a potentially remarkable computational speed up.

Due to the limitations of supervised methods, a number of works have then considered so called \emph{unsupervised} approaches \cite{balakrishnan2019voxelmorph, deVos2019deep} where a neural network is trained based on the original cost function of traditional intensity-based methods minimising a dissimilarity measure such as mean squared intensity differences and others. While training over large number of examples might indeed be beneficial for optimising the cost function (it could have a regularisation effect or improve generalisation), these unsupervised approaches cannot be expected to perform fundamentally better compared to traditional methods as the exact same function is optimised. One might even argue that traditional methods are more flexible, as they can adapt to any new pair of test images, and are not limited to register images of similar appearance as the training data. For example, many of the traditional methods discussed in \cite{sotiras2013deformable} can be equally used for brain MRI and lung CT with maybe only a few changes to some hyper-parameters. A recent hybrid approach \cite{fan2019birnet} is combining an unsupervised and supervised cost function using a traditional method to generate training deformation fields.

\begin{figure}[t]
    \centering
    \includegraphics[width=\linewidth]{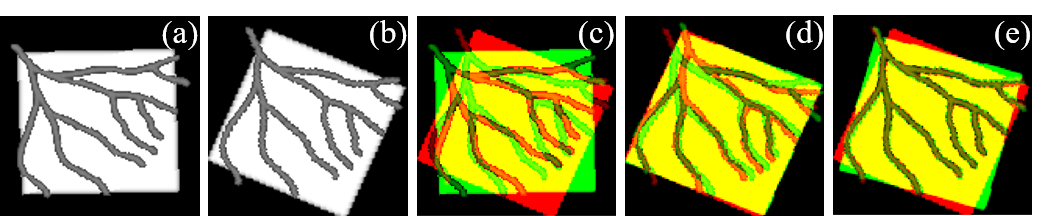}
    \caption{A toy example (vessel trees on white box) illustrating the benefit of structure-guided image registration. The initial alignment of images (a,b) is shown in (c) and after intensity-based affine registration in (d). Registration focuses on aligning the white box ignoring \acrfull{soi}, i.e., the vessels. Our ISTNs learn to focus on the \acrshort{soi} yielding accurate alignment in (e).}
    \label{fig:illustration}
\end{figure}

Overall, one may argue that neural network based image registration, so far, has not taken full advantage of deep representation learning but mostly led to a speed up in the time it takes to register two images. At the same time, one may argue that registration accuracy is of higher importance than speed in many clinical applications. We observe that neither the supervised nor the unsupervised methods exploit two key advantages of neural networks, which are 1) the ability to learn new representations that are optimised for a downstream task, and 2) the ability to incorporate and benefit from additional information during training that is unavailable (or very difficult to obtain) at test time.

Some exceptions to the second point are works that consider extra information such as segmentations or weak labels during training \cite{rohe2017svf, hu2018weakly, balakrishnan2019voxelmorph}. This additional supervision can help to guide the registration at test time in a different way than using image intensities alone. For example, the registration may focus on particular \acrshort{soi} (cf. Fig~\ref{fig:illustration}). However, the current approaches do not retain or explicitly extract such extra information, so it cannot be used further at test time, for example, for refining the predicted, initial transformation parameters. In fact, most of the current works consider neural network based registration as a one-pass (non-iterative) process, which might be sub-optimal as we show in our results. The few works that discuss subsequent refinement either suggest to use a traditional (iterative) method \cite{balakrishnan2019voxelmorph}, or to use the network in an auto-regressive way \cite{deVos2019deep}. As both rely again on optimising the original intensity-based cost function the advantage over a traditional method remains unclear, and any extra information that was available during training is somewhat lost.

\begin{figure}[t]
    \centering
    \includegraphics[width=1.0\linewidth]{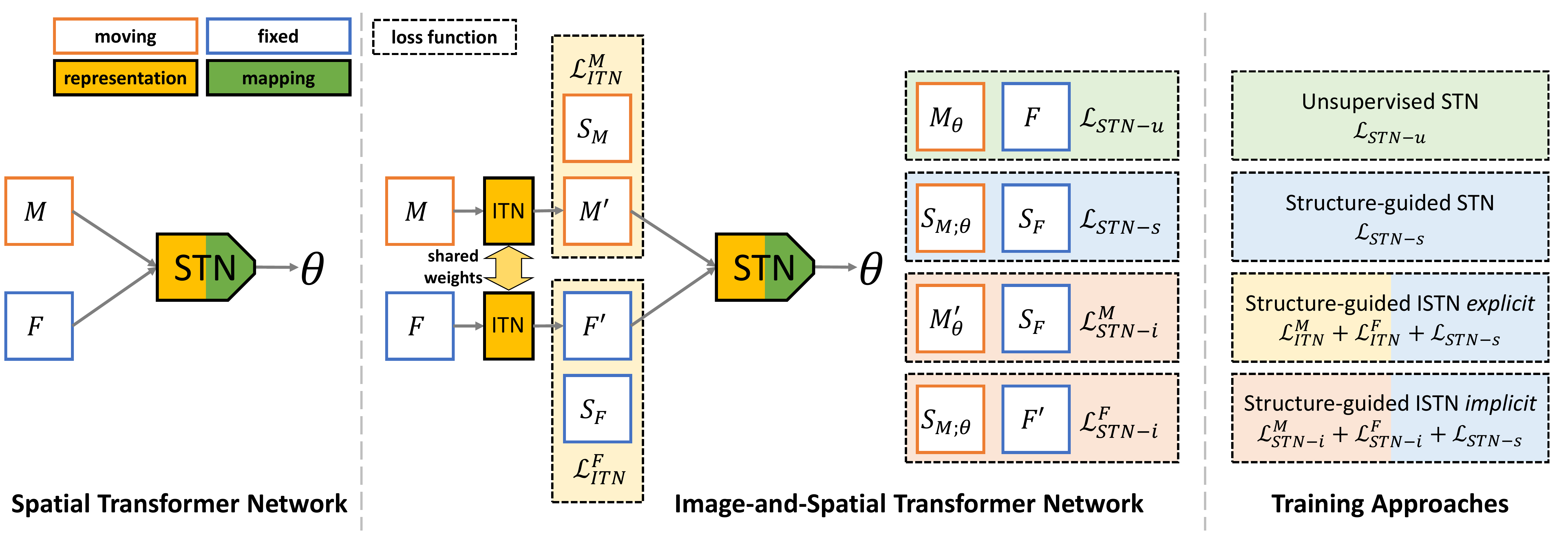}
    \caption{Overview of image registration using \acrfull{stn}, with the classical model shown on the left mapping an image pair $M,F$ \emph{directly} to parameters $\theta$ of a spatial transformation (cf. \cite{balakrishnan2019voxelmorph,deVos2019deep}). Our \acrfull{istn} introduce a dedicated image transformer network (ITN) to learn to produce image representations $M',F'$ optimised for the downstream registration task, as well as predicting $\theta$. This gives raise to multiple ways of training \acrshort{istn} by combining different loss functions (see Sec~\ref{sec:training}).}
    \label{fig:overview}
\end{figure}

\subsection{Contributions}
To overcome these limitations, and to make best use of the key ability of neural networks to learn representations, we introduce \acrfull{istn} where a dedicated Image Transformer Network (ITN) is added to the head of a Spatial Transformer Network (STN) aiming to extract and retain information about \acrshort{soi}s, for which annotations are only required during training. While the STN predicts the parameters of the spatial transformation, the ITN produces a new image representation which is learned in an end-to-end fashion and optimised for the downstream registration task. This allows us to not only predict a good initial transformation at test time, but enables what we call \emph{structure-guided} registration with an accurate test-specific, iterative refinement using the exact same model. An illustrative example of what ISTNs can do and why structure-guided registration can be useful is shown in Fig.~\ref{fig:illustration}. A schematic overview of our approach and how it relates to previous work that uses STNs only (such as \cite{balakrishnan2019voxelmorph,deVos2019deep}) is shown in Fig.~\ref{fig:overview}.

\section{Image-and-Spatial Transformer Networks}
\label{sec:method}

Spatial Transformer Networks \cite{jaderberg2015spatial} are the building block of most of the recent works on neural network based image registration. An STN is a neural network in itself commonly consisting of a few convolutional and fully connected layers that are able to learn a mapping from input images $(M, F)$ to parameters $\theta$ of a spatial transformation $T_{\theta}$. Taking advantage of the fact that image re-sampling is a differentiable operation, STNs can be trained end-to-end, and plugged as a module into larger networks (as originally used for improving image classification~\cite{jaderberg2015spatial}). Revisiting the structure of an STN, we observe that there are two main components: a feature extraction part learning a new representation of the input using convolutional layers, and a second part that maps these representations to transformation parameters. We indicate this in Fig.~\ref{fig:overview} using different colours within the STN module. The representation that STNs may learn, however, is not exposed and remains hidden during inference. This is where our main contribution comes into play where we redesign the basic building block of the transformer module of neural network based image registration by introducing a dedicated Image Transformer Network.



\subsection{Image Transformer Networks} 
We define ITNs to be convolutional neural networks that map an input image to an output image of the same size and dimension. In this paper we consider the case where the number of channels is the same for the input and output, although this does not have to be the case and other variants may be considered. The role of the ITN is to expose explicitly a learned image representation that is optimal for the downstream registration task solved by the STN. A shared ITN for inputs $M$ and $F$ learns to generate outputs $M'$ and $F'$ which are fed into a regular STN (cf. Fig.~\ref{fig:overview}).
This new architecture gives raise to a number of training approaches, in particular, when extra information about SoIs are available, such as image segmentations or landmarks.



\begin{figure}[t]
    \centering
    \includegraphics[width=\linewidth]{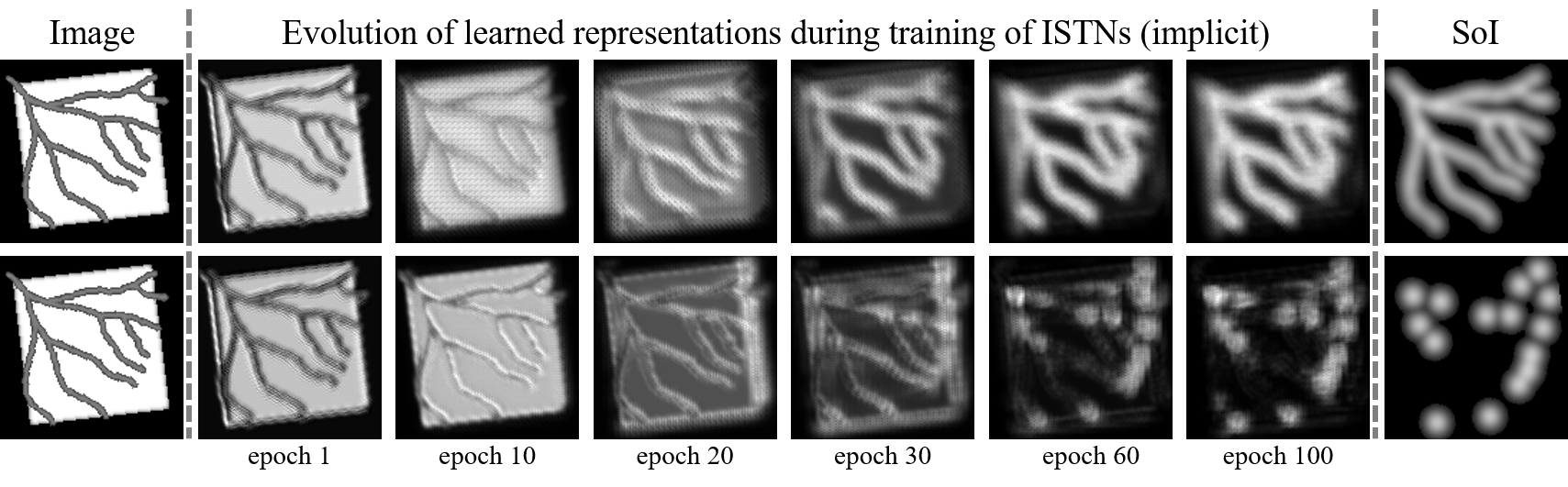}
    \caption{For a toy example, we show the progress of the output of the ITN module in an ISTN model trained with the \emph{implicit} loss function $\mathcal{L}_{ISTN-i}$ (cf. Eq.~(\ref{eq:istn-i}) in Sec.~\ref{sec:training} and Fig.~\ref{fig:overview}). The top row corresponds to a case where the SoI is a segmentation-like mask (shown on the most right). The bottom row shows training with landmark maps. The learned image representation allows accurate structure-guided registration with test-specific refinement at test time.}
    \label{fig:reps-implicit}
\end{figure}

\subsection{Explicit and Implicit Training of ISTNs}
\label{sec:training}
As indicated in Fig.~\ref{fig:overview}, different loss functions can be considered for training ISTNs. Note that the unsupervised case using image intensities only, here corresponding to $\mathcal{L}_{STN-u}(M_{\theta},F)$, is a special case of training an ISTN. We use $M_{\theta}$ as a short form of $M \circ T_{\theta}$. Similarly, we can incorporate auxiliary information in form of segmentations as in \cite{balakrishnan2019voxelmorph} or other structural or geometric information via a `structure-guided' or `supervised' loss $\mathcal{L}_{STN-s}(S_{M;\theta},S_F)$. Here, $S$ are images encoding SoIs (e.g., organ segmentations, anatomical landmarks, centerlines, etc.), and `supervised' refers to the fact that such SoIs need annotations on the training data. Note, that neither $\mathcal{L}_{STN-u}$ nor $\mathcal{L}_{STN-s}$ will encourage the ITN to learn any directly useful representations. In order to facilitate this, we propose two different strategies for training ISTNs when auxiliary information about SoIs is available. The loss function for our \emph{explicit} ISTN is defined as
\begin{equation}
\label{eq:istn-e}
\mathcal{L}_{ISTN-e} = \mathcal{L}^{M}_{ITN} + \mathcal{L}^{F}_{ITN} + \mathcal{L}_{STN-s}
\end{equation}
which combines the supervised STN with an ITN loss $\mathcal{L}_{ITN}(S_I,I')$ explicitly penalizing differences between the SoI encoding $S_I$ and the ITN output $I'$ for input image $I$. While the explicit loss has the desired effect of producing representations capturing the SoI information, the ITN and STN losses are somewhat decoupled where the ITN loss plays the role of deep supervison.

An intriguing alternative is the \emph{implicit} ISTN with a loss function defined as
\begin{equation}
\label{eq:istn-i}
\mathcal{L}_{ISTN-i} = \mathcal{L}^{M}_{STN-i} + \mathcal{L}^{F}_{STN-i} + \mathcal{L}_{STN-s}
\end{equation}
Here, the two terms $\mathcal{L}^{M}_{STN-i}(M'_{\theta},S_F)$ and $\mathcal{L}^{F}_{STN-i}(S_{M;\theta},F')$ intertwine the outputs of the ITN $(M',F')$, the SoI encodings $(S_M,S_F)$ and the estimated transformation parameters $\theta$ from the STN. Combined with the supervised STN loss this gives raise to a fully end-to-end training of image representations that are optimised for the downstream registration task. In Fig.~\ref{fig:reps-implicit} we show how the ITN representations evolve over the course of training on an illustrative toy example both for the case of segmentations and landmark annotations. Due to space reasons, we omit the figure for the explicit ISTN which shows similar results with slightly sharper representations due to the ITN loss.

In this paper, for all above mentioned loss functions we use the mean squared error (MSE) loss. Other losses can be considered but we find MSE to work very well. It also allows us to flexibly incorporate different types of SoI information by simply representing it in the form of real-valued images. This is straightforward for binary segmentations, and anatomical landmarks can be, for example, encoded via distance maps or smoothed centroid maps (cf. bottom row of Fig.~\ref{fig:reps-implicit}).


\subsection{Test-specific Iterative Refinement}

The key of ISTNs is that they enable structure-guided, test-specific refinement based on the learned representations $M'$ and $F'$ that are exposed by the ITN. Though any registration technique can be used for refinement by using the inferred images $M'$ and $F'$ as inputs, we can also directly leverage the STN module itself to perform the refinement at inference time by iteratively updating the STN weights (keeping the ITN weights fixed) through minimization of a refinement loss denoted as $\mathcal{L}_{STN-r}(M'_{\theta},F')$. Note, that no annotations are required for the test images to perform this iterative refinement, as the necessary representations $M'$ and $F'$ are generated by the trained ITN.


\subsection{Transformation Models}
\label{sec:transformation}

STNs make no assumption about the employed spatial transformation model, and hence, our ISTN architecture remains generic allowing the integration of various linear and non-linear transformation models. In this paper, we consider two specific transformation models as a proof-of-concept. We employ a typical parameterisation for affine transformations decomposed into
\begin{equation}
T^{\textrm{affine}}_{\theta} = M_t R_\phi S^{-1}_\psi D_s S_\psi
\end{equation}
where $M_t$, $R_\phi$, $D_s$ and $S_\psi$ are the translation, rotation, scaling and shearing matrices. This decomposition allows us to set intuitive bounds on the individual transformation parameters by using appropriately scaled $\tanh$ functions.
For non-linear registration we employ a standard B-spline parameterisation using a regular grid of control points. Transformation parameters $\theta$ then correspond to control point displacements. Details on this, with emphasis on their applications and implementations within neural network architectures can be found in \cite{deVos2019deep,sandkuhler2018airlab}.



\section{Experiments}
\label{sec:experiments}

We evaluate ISTNs on the task of pairwise registration of brain MRI (similar to \cite{balakrishnan2019voxelmorph,fan2019birnet}). In the first set of experiments, we use 420 individual subjects from the UK Biobank Imaging Study\footnote{UK Biobank Resource under Application Number 12579} to form 100 random pairs of moving and fixed images for training, 10 pairs for validation, and 100 pairs for testing. For each image, segmentations of sub-cortical structures are available. Images are skull-stripped and intensity normalised. Binarised segmentation label maps are used as the SoI information. Due to space reasons, we focus on the results while our publicly available code\footnote{\url{https://github.com/biomedia-mira/istn}} contains details on architectural choices, hyper-parameters and training configurations.

We assess agreement of SoI after registration by calculating Dice scores and average surface distances (ASDs). We compare the explicit and implicit variants (ISTN-e and ISTN-i) with unsupervised and supervised models (STN-u and STN-s). These baselines are conceptually similar to the approaches in \cite{deVos2019deep,balakrishnan2019voxelmorph}. The baselines have been set up in a competitive way and we checked for proper convergence and best possible performance on the validation set.

\begin{figure}[t]
    \centering
    \includegraphics[width=0.13\linewidth]{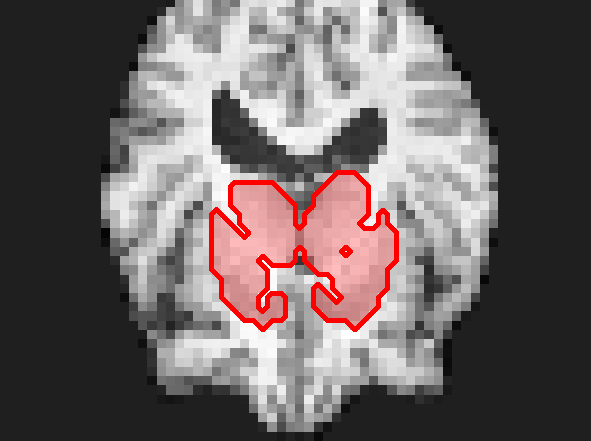}
    \includegraphics[width=0.13\linewidth]{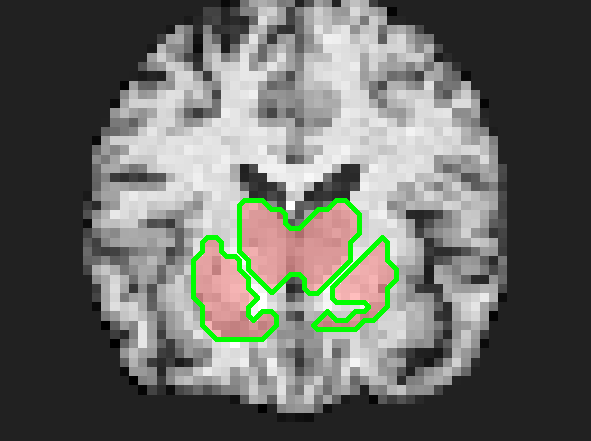}
    \includegraphics[width=0.13\linewidth]{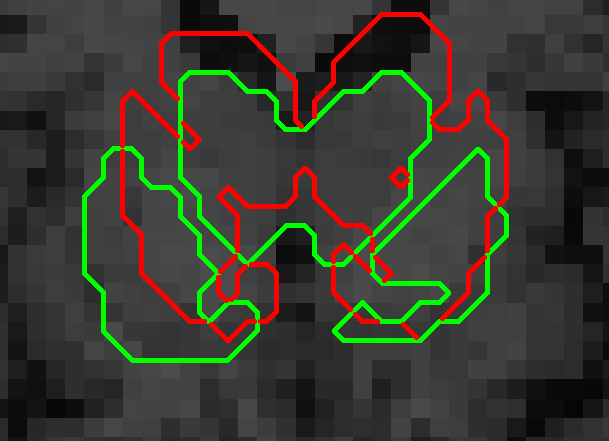}
    \includegraphics[width=0.13\linewidth]{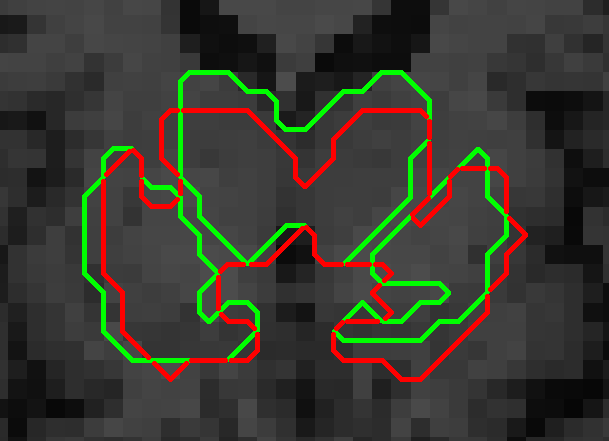}
    \includegraphics[width=0.13\linewidth]{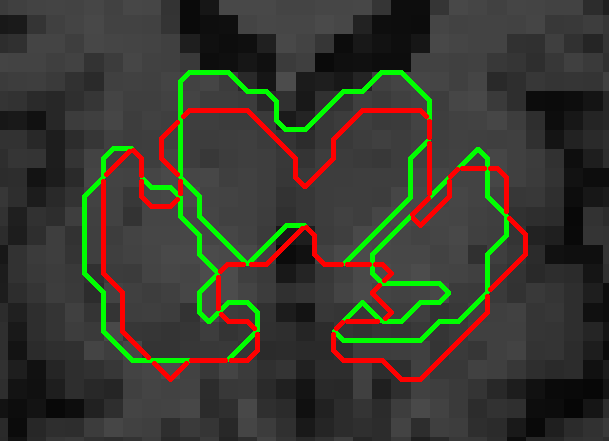}
    \includegraphics[width=0.13\linewidth]{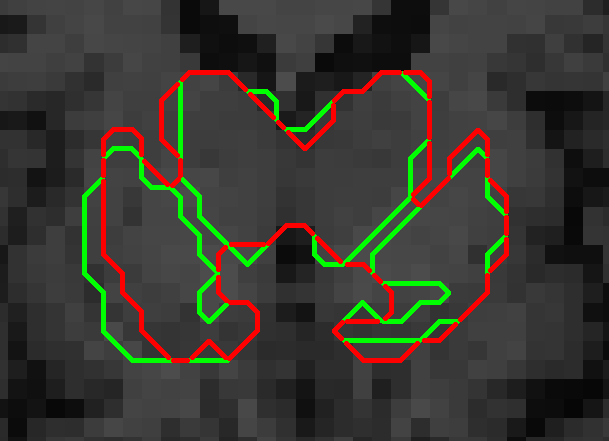}
    \includegraphics[width=0.13\linewidth]{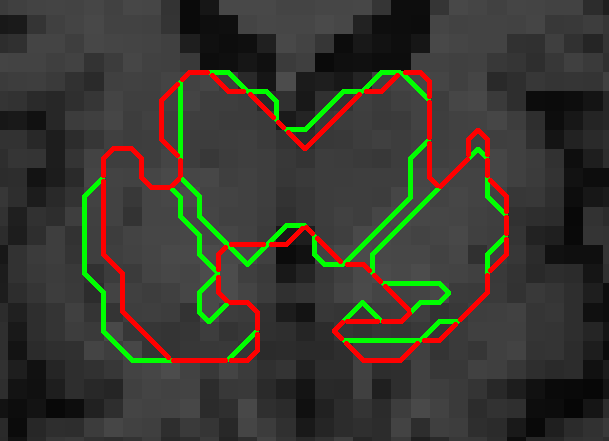}
    \includegraphics[width=0.13\linewidth]{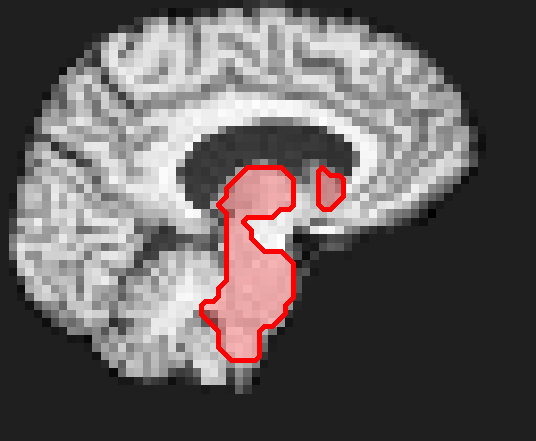}
    \includegraphics[width=0.13\linewidth]{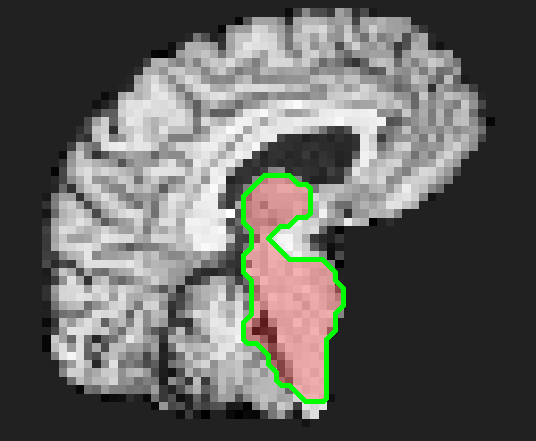}
    \includegraphics[width=0.13\linewidth]{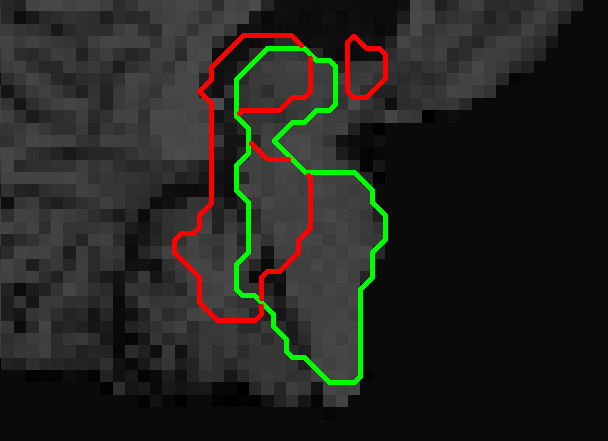}
    \includegraphics[width=0.13\linewidth]{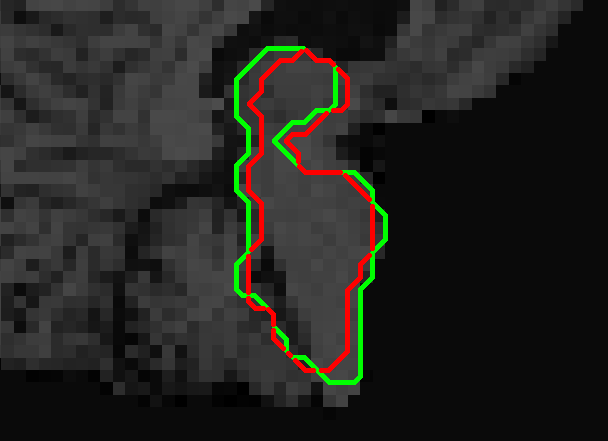}
    \includegraphics[width=0.13\linewidth]{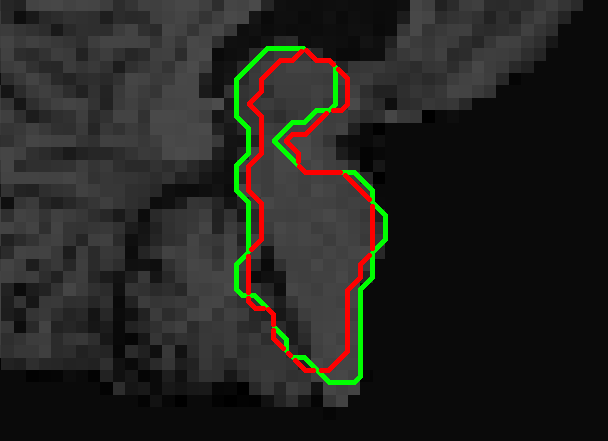}
    \includegraphics[width=0.13\linewidth]{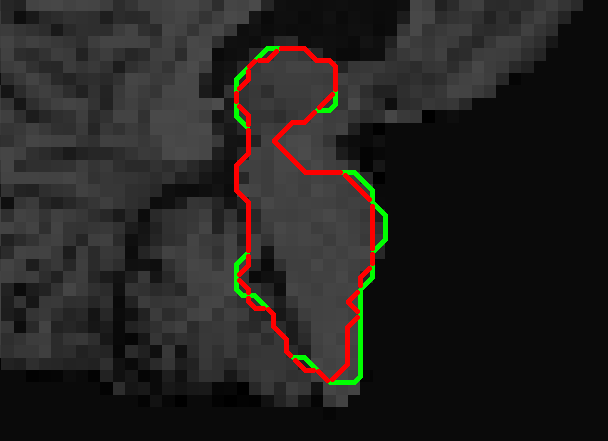}
    \includegraphics[width=0.13\linewidth]{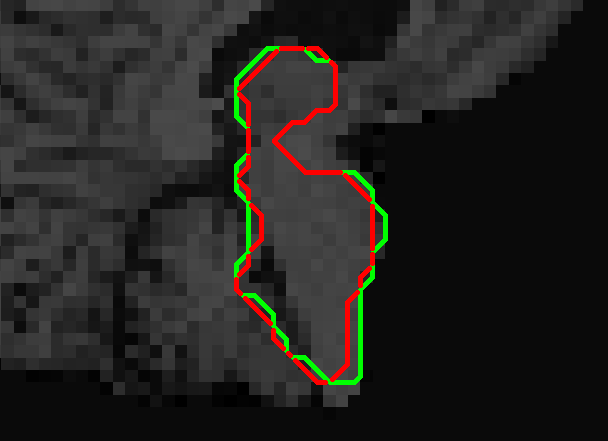}
    \caption{Visual results for affine registration with SoI overlaid in red (moving), and green (fixed). On the left: the input images before registration. Close-ups from left to right: initial alignment, STN-u, STN-s, ISTN-e, ISTN-i.}
    \label{fig:brain}
\end{figure}

\begin{figure}[t]
    \centering
    \includegraphics[width=0.16\linewidth]{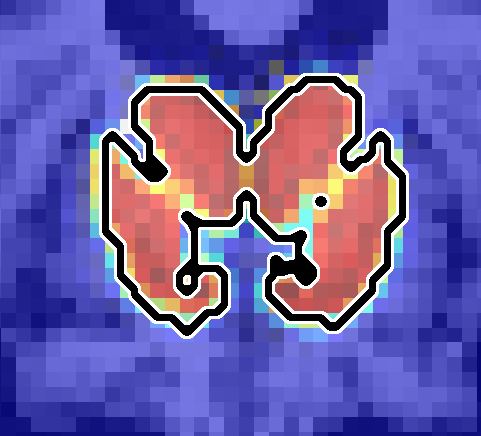}
    \includegraphics[width=0.16\linewidth]{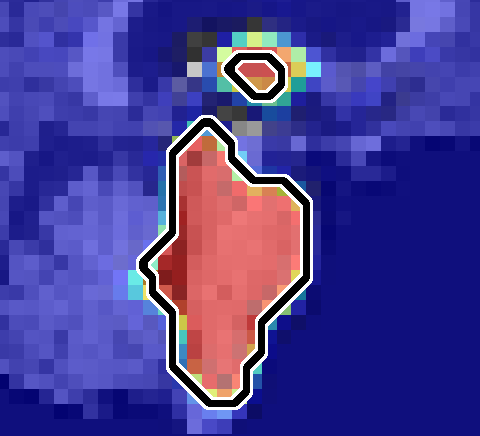}
    \includegraphics[width=0.16\linewidth]{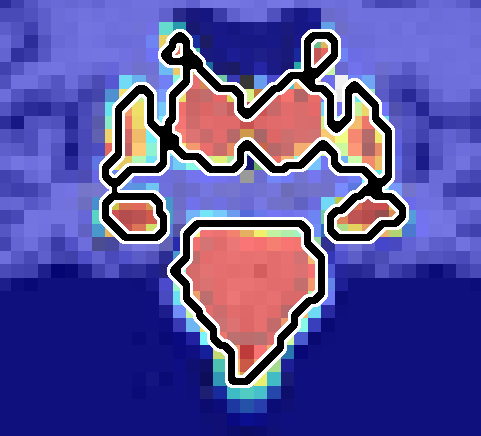}
    \includegraphics[width=0.16\linewidth]{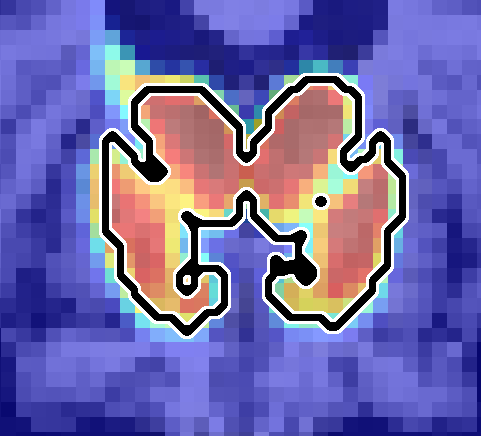}
    \includegraphics[width=0.16\linewidth]{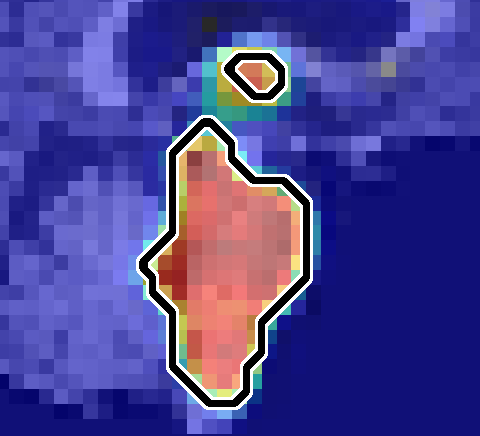}
    \includegraphics[width=0.16\linewidth]{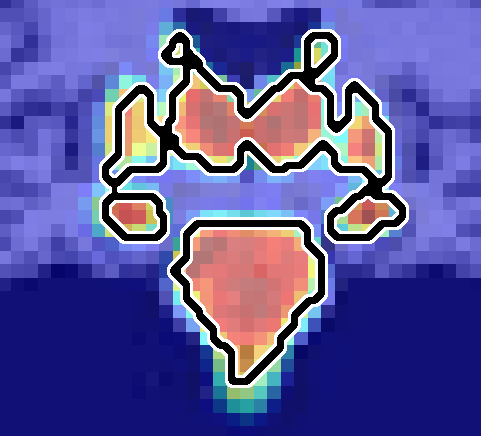}
    \caption{Overlay of the learned representations of sub-cortical structures as heatmaps on top of a test input. The axial, sagittal and coronal slices on the left correspond to the ITN output of an \emph{explicit} ISTN-e, the left shows the output for an \emph{implicit} ISTN-i. Black contours show the ground truth segmentations.}
    \label{fig:prime}
\end{figure}

Table~\ref{tab:results_affine_100} summarizes the quantitative results before and after iterative refinement using the four different approaches with affine and non-linear B-spline transformation models. For the B-splines we use a 30 mm control point spacing and start registration from rigidly pre-aligned images. For both affine and B-splines we provide numbers for the initial state (Id) and results when the SoI is used directly for registration as an upper bound ``best-case'' reference. We note that ISTN-e/i outperform STN-u/s for the one-pass predictions. Test-specific refinement boosts the accuracy significantly for all approaches. ISTNs achieve overall highest Dice scores and lowest ASDs, sometimes close to the best-case. STN-u/s after refinement converge to the same lower accuracy as no SoI information can be leveraged. Fig.~\ref{fig:brain} shows qualitative results after refinement for affine registration, while Fig.~\ref{fig:prime} shows an example of learned SoI representations.

We repeat the experiments but with 1,000 pairs for training instead of 100. The one-pass predictions improve for all four methods with Dice scores of 0.75 (STN-u), 0.77 (STN-s), and 0.80 (ISTN-e/i) for affine, and 0.79, 0.83, and 0.86 (same order) for non-linear. Iterative refinement yields quasi identical results for STN-u/s as before when trained with 100 pairs, and slightly improves for ISTNs. Remarkably, ISTNs trained with only 100 pairs achieve much higher accuracy than STN-u and STN-s trained with 1,000 pairs, indicating excellent data efficiency and benefit of iterative, test-specific refinement for image registration.
{
\begin{table}[t]
\begin{center}
\caption{Summary of registration results when using 100 training images.}
\label{tab:results_affine_100}
\begin{tabular}{ccccccccc}
\toprule
& &  &  & \multicolumn{4}{c}{\cellcolor{lightgray}Evaluated Methods} &  \\
$T_{\theta}$&Metric & Refine & Id & STN-u & STN-s & ISTN-e & ISTN-i & SoI \\
\midrule
\multirow{4}{*}{\STAB{\rotatebox[origin=c]{90}{Affine}}}&\multirow{2}{*}{Dice} & before & \multirow{2}{*}{0.53} & 0.71 & 0.70 & 0.75 & 0.75 & \multirow{2}{*}{0.84} \\
& & after & & 0.79 & 0.79 & 0.83 & 0.82 & \\
\cmidrule{2-9}
&\multirow{2}{*}{ASD} & before & \multirow{2}{*}{2.41} & 1.11 & 1.10 & 0.93 & 0.89 & \multirow{2}{*}{0.50} \\
& & after & & 0.69 & 0.69 & 0.53 & 0.58 & \\

\hdashline
\multirow{4}{*}{\STAB{\rotatebox[origin=c]{90}{B-Spline}}}&\multirow{2}{*}{Dice} & before & \multirow{2}{*}{0.70} & 0.74 & 0.77 & 0.80 & 0.81 & \multirow{2}{*}{0.91} \\
& & after & & 0.84 & 0.83 & 0.86 & 0.85 & \\
\cmidrule{2-9}
&\multirow{2}{*}{ASD} & before & \multirow{2}{*}{1.05} & 0.88 & 0.74 & 0.63 & 0.59 & \multirow{2}{*}{0.27} \\
& & after & & 0.51 & 0.52 & 0.45 & 0.46 & \\
 \bottomrule
\end{tabular}
\end{center}
\end{table}
}

\section{Conclusion}
\label{sec:conclusion}

ISTNs are a generic framework for neural network based structure-guided image registration with test-specific refinement using learned representations. In our experiments the explicit and implicit variants perform equally well and outperform unsupervised and supervised STNs both before and after refinement. Implicitly learned representations may be beneficial to prevent overfitting in cases where SoI information contains noise or corruption. This effect and other applications of ISTNs using different types of SoI will be explored in future work.

\bibliographystyle{splncs}


\end{document}